\documentclass[11pt,a4paper]{article}
\usepackage[hyperref]{acl2017}
\usepackage{times}
\usepackage{latexsym}

\usepackage{url}

\usepackage{adjustbox,lipsum}
\usepackage{color}

\aclfinalcopy % Uncomment this line for the final submission
\title{Dependency Language Models for Transition-based Dependency Parsing}

\author{Juntao Yu \\
  University of Birmingham \\
  Birmingham, UK \\
  {\tt j.yu.1@cs.bham.ac.uk} \\\And
  Bernd Bohnet \\
  Google\\
  London, UK \\
 {\tt bohnetbd@google.com} \\}

\date{}

\begin{document}
\maketitle
\begin{abstract}
In this paper, we present an approach to improve the accuracy of a strong transition-based dependency parser by exploiting dependency language models that are extracted 
from a large parsed corpus. We integrated a small number of features based on the dependency language models into the parser.   To demonstrate the effectiveness of the proposed approach, we evaluate our parser on standard English and Chinese data where the base parser could achieve competitive accuracy scores. Our enhanced parser achieved state-of-the-art accuracy on Chinese data and competitive results on English data. We gained a large absolute improvement of one point (UAS) on Chinese and 0.5 points for English. 
\end{abstract}

\section{Introduction}
In recent years, using unlabeled data to improve natural language parsing has seen a surge of interest as the data can easy and inexpensively be obtained, cf. \cite{sarkar01,steedman03,mcclosky06naacl,koo08,sogaard2010,petrov2012,chen2013feature,weiss-etAl:2015:ACL}. This is in stark contrast to the high costs of manually labeling new data. Some of the techniques such as self-training \cite{mcclosky06naacl} and co-training \cite{sarkar01} use auto-parsed data as additional training data. This enables the parser to learn from its own or other parser's annotations. Other techniques include word clustering \cite{koo08} and word embedding \cite{Bengio:2003:NPL:944919.944966} which are generated from a large amount of unannotated data. The outputs can be used as features or inputs for parsers. Both groups of techniques have been shown effective on syntactic parsing tasks \cite{zhou2005tri,reichart2007self,sagae2010self,sogaard2010,yu-elkaref-bohnet:2015:IWPT,weiss-etAl:2015:ACL}.  However, most word clustering and the word embedding approaches do not consider the syntactic structures and most self-/co-training approaches can use only a relatively small additional training data as training parsers on a large corpus might be time-consuming or even intractable on a corpus of millions of sentences. 

Dependency language models (DLM) \cite{shen2008new}  are variants of language models based on dependency structures. An N-gram DLM is able to predict the next child when given N-1 immediate previous children and their head.  \newcite{chen2012utilizing} integrated first a high-order DLM into a second-order graph-based parser. The DLM allows the parser to explore high-order features but not increasing the time complexity. 
Following \newcite{chen2012utilizing}, we adapted the DLM to transition-based dependency parsing. Our approach is different from \newcite{chen2012utilizing}'s in a number of important aspects: 

\begin{enumerate}
	\item We applied the DLM to a strong parser that on its own has a competitive performance. \item We revised their feature templates to integrate the DLMs with a transition-based system and labeled parsing. 
	\item We used DLMs in joint tagging and parsing, and gained up to 0.4\% on tagging accuracy.
	\item Our approach could use not only single DLM but also multiple DLMs during parsing. \item We evaluated additionally with DLMs extracted from higher quality parsed data which two parsers assigned the same annotations.
\end{enumerate}

Overall, our approach improved upon a competitive baseline by 0.51\% for English and achieved state-of-the-art accuracy for Chinese.

\section{Related work}
\label{relwork}
Previous studies using unlabeled text could be classified into two groups by how unlabeled data is used for training. 

The first group uses unlabeled data (usually parsed data) directly in the training process as additional training data. The most common approaches in this group are self-/co-training.  \newcite{mcclosky06naacl} applied first self-training to a constituency parser. This was later adapted to dependency parsing by \newcite{kawahara2008learning} and \newcite{yu-elkaref-bohnet:2015:IWPT}. 
Compared to the self-training approach used by \newcite{mcclosky06naacl}, both self-training approaches for dependency parsing need an additional 
selection step to predict high-quality parsed sentences for retraining. The basic idea behind this is similar to \newcite{sagae07}'s co-training approach. Instead of using a separately trained classifier \cite{kawahara2008learning} or confidence-based methods \cite{yu-elkaref-bohnet:2015:IWPT}, \newcite{sagae07} used two different parsers to obtain the additional training data. \newcite{sagae07} shows that when two parsers assign the same syntactic analysis to sentences then the parse trees have usually a higher parsing accuracy. Tri-training \cite{zhou2005tri,sogaard2010} is a variant of co-training which involves a third parser. The base parser is retrained on additional parse trees that the other two parsers agreed on. 

The second group uses the unlabeled data indirectly. \newcite{koo08} used word clusters built from unlabeled data to train a parser. \newcite{chen2008learning} used features extracted from short distance relations of a parsed corpus to improve a dependency parsing model. \newcite{suzuki-EtAl:2009:EMNLP} used features of generative models estimated from large unlabelled data to improve a second order dependency parser. Their enhanced models improved upon the second order baseline models by 0.65\% and 0.15\% for English and Czech respectively. \newcite{Mirroshandel12} used the relative frequencies of nine manually selected head-dependent patterns calculated from parsed French corpora to rescore the n-best parses.
Their approach gained a labeled improvement of 0.8\% over the baseline. \newcite{chen2013feature} combined meta features based on frequencies with the basic first-/second-order features. 
The meta features are extracted from parsed annotations by counting the frequencies of basic feature representations in a large corpus. 
With the help of meta features, the parser achieved the state-of-the-art accuracy on Chinese. \newcite{kiperwasser-goldberg:2015:EMNLP} added features based on the statistics learned from unlabeled data to a weak first-order parser and they achieved 0.7\% improvement on the English data. Word embeddings that represent words as high dimensional vectors are mostly used in neural network parsers \cite{chen2014fast,weiss-etAl:2015:ACL} and play an important role in those parsers. The approach most close to ours is reported by \newcite{chen2012utilizing} who applied a high-order DLM to a second-order graph-based parser for unlabeled parsing. 
Their DLMs are extracted from an English corpus that contains 43 million words \cite{charniak00} and a 311 million word corpus of Chinese \cite{huang09} parsed by a parser. From a relatively weak baseline, additional DLM-based features gained 0.6\% UAS for English and an impressive 2.9\% for Chinese.

\section{Our Approach}
\label{approach}

Dependency language models were introduced by \newcite{shen2008new} to capture long distance relations in syntactic structures. 
An N-gram DLM predicts the next child based on N-1 immediate previous children and their head. 
We integrate DLMs extracted from a large parsed corpus into the Mate parser \cite{bohnet2013joint}. We first extract DLMs from a corpus parsed by the base model. We then retrain the parser with additional DLM-based features.

Further, we experimented with techniques to improve the quality of the syntactic annotations which we use to build the DLMs. 
We parse the sentences with two different parsers and then select the annotations which both parsers agree on. 
The method is similar to co-training except that we do not train the parser directly on these sentences.

We build the DLMs with the method of \newcite{chen2012utilizing}. 
For each child $x_{ch}$, we gain the probability distribution $P_u(x_{ch}|H)$, where $H$ refers $N-1$ immediate previous children and their head $x_h$. The previous children for $x_{ch}$ are those who share the same head with $x_{ch}$ but closer to the head word according to the word sequence in the sentence. Let's consider the left side child $x_{Lk}$ in the dependency relations $(x_{Lk}...x_{L1}, x_h, x_{R1}...x_{Rm})$ as an example, the N-1 immediate previous children for $x_{Lk}$ are $x_{Lk-1}..x_{Lk-N+1}$.  
In our approach, we estimate $P_u(x_{ch}|H)$ by the relative frequency: %$P_u(x_{ch}|H) = \frac{count(x_{ch},H)}{\sum_{x'_{ch}} count(x'_{ch},H)}$
\begin{equation}
P_u(x_{ch}|H) = \frac{count(x_{ch},H)}{\sum_{x'_{ch}} count(x'_{ch},H)}
%\vspace{0.3cm}
\end{equation}

By their probabilities, the N-grams are sorted in a descending order. We then used the thresholds of \newcite{chen2012utilizing} to replace the probabilities with one of the three classes ($PH,PM,PL)$ according to their position in the sorted list, i.e. the N-grams whose probability has a rank in the first 10\% receives the tag $PH$, $PM$ refers probabilities ranked between 10\% and 30\%, probabilities that ranked below 30\% are replaced with $PL$. 
During parsing, we use an additional class $PO$ for relations not presented in the DLM. In the preliminary experiments, the $PH$ class is mainly filled by unusual relations that only appeared a few times in the parsed text. To avoid this, we configured the DLMs to only use elements which have a minimum frequency of three, i.e. $count(x_{ch},H) \geq 3$.
Table \ref{table-feature} shows our feature templates, where $NO_{DLM}$ is an index which allows DLMs distinguish from each other, $s_0$, $s_1$ are the top and the second top of the stack, $\phi(P_u(s_0/s_1))$ refers the coarse label of probabilities $P_u(x_{s_0/s_1}|H)$ (one of the $PH,PM,PL,PO$), 
$s_0/s_1\_pos, s_0/s_1\_word$ refer to the part-of-speech tag, word form of $s_0/s_1$, and $label$ is the dependency label between the $s_0$ and the $s_1$.

\begin{table}[t]
\begin{center}
\begin{adjustbox}{max width=7cm}
\begin{tabular}{l}
\hline  \( <NO_{DLM}, \phi(P_u(s_0)), \phi(P_u(s_1)), label> \) \\
\( <NO_{DLM}, \phi(P_u(s_0)), \phi(P_u(s_1)), label, s_0\_pos> \) \\
\( <NO_{DLM}, \phi(P_u(s_0)), \phi(P_u(s_1)), label, s_0\_word> \) \\
\( <NO_{DLM}, \phi(P_u(s_0)), \phi(P_u(s_1)), label, s_1\_pos> \) \\
\( <NO_{DLM}, \phi(P_u(s_0)), \phi(P_u(s_1)), label, s_1\_word> \) \\
\( <NO_{DLM}, \phi(P_u(s_0)), \phi(P_u(s_1)), label, s_0\_pos, s_1\_pos> \) \\
\( <NO_{DLM}, \phi(P_u(s_0)), \phi(P_u(s_1)), label, s_0\_word, s_1\_word> \) \\
 \hline

\end{tabular}
\end{adjustbox}
\end{center}
\caption{\label{table-feature} Feature templates which we use in the parser.}
\end{table}

\section{Experimental Set-up}
\label{setup}

\begin{table}[t]
\begin{center}
\begin{adjustbox}{max width=6cm}
\begin{tabular}{l|l|l|l}
\hline  &\bf train&\bf dev &\bf test \\ \hline
PTB&  2-21 &22 &23\\ \hline
CTB5& 001-815, & 886-931, &816-885, \\
 &1001-1136&1148-1151&1137-1147\\
 \hline

\end{tabular}
\end{adjustbox}
\end{center}
\caption{\label{table-datasplits} Our data splits for English and Chinese}
\end{table} 

For our experiments, we used the Penn English Treebank (PTB) \cite{marcus93} and Chinese Treebank 5 (CTB5) \cite{xue05}.
For English, we follow the standard splits and 
%section 2-21 are used for training, section 22 is used as the development set and 23 as the final test set.
used Stanford parser
\footnote{http://nlp.stanford.edu/software/lex-parser.shtml} 
v3.3.0 to convert the constituency trees into Stanford style dependencies \cite{demarneffe06}. For Chinese, we follow the splits of \newcite{zhang11}, the constituency trees are converted to dependency relations by Penn2Malt\footnote{http://stp.lingfil.uu.se/~nivre/research/Penn2Malt.html} tool using head rules of \newcite{zhang08}. Table \ref{table-datasplits} shows the splits of our data. We used gold segmentation for Chinese tests to make our work comparable with previous work. We used predicted part-of-speech tags for both languages in all evaluations. Tags are assigned by base parser's internal joint tagger trained on the training set. We report labeled  (LAS) and unlabeled (UAS) attachment scores, punctuation marks are excluded from the evaluation. 

For the English unlabeled data, we used the data of \newcite{Chelba13onebillion} which contains around 30 million sentences (800 million words) from the news domain. For Chinese, we used Xinhua portion of Chinese Gigaword
\footnote{We excluded the sentences of CTB5 from Chinese Gigaword corpus.} 
Version 5.0 (LDC2011T13). The Chinese unlabeled data we used consists of 20 million sentences which is roughly 450 million words after being segmented by ZPar\footnote{https://github.com/frcchang/zpar} v0.7.5. The word segmentor is trained on the CTB5 training set. In most of our experiments, the DLMs are extracted from data annotated by our base parser. For the evaluation on higher quality DLMs, the unlabeled data is additionally tagged and parsed by Berkeley parser \cite{petrov07} and is converted to dependency trees with the same tools as for gold data.

We used Mate transition-based parser with its default setting and a beam of 40 as our baseline. 

\section{Results and Discussion}
\label{results}

%\subsection{Development Experiments}

%{\color{red} {\em We explore adding up to N-gram} seems difficult to understand. Can you rewrite? Can you provide a distinct number(s) for N?}
\textbf{Combining different N-gram DLMs.} We first evaluated the effects of adding different number of DLMs. Let $m$ be the DLMs we used in the experiments, e.g. $m$=1-3 refers all three (unigram, bigram and trigram) DLMs are used. We evaluate with both single and multiple DLMs that extracted from 5 million sentences for both languages. 
We started from only using unigram DLM ($m$=1) and then increasing the $m$ until the accuracy drops.   
Table \ref{table-order} shows the results with different DLM settings. The unigram DLM is most effective for English, which improves above the baseline by 0.38\%. For Chinese, our approach gained a large improvement of 1.16\% with an $m$ of 1-3. Thus, we use $m$=1 for English and $m$=1-3 for Chinese in the rest of our experiments.

%the influence of DLMs extracted from higher quality data
\textbf{Exploring DLMs built from corpora of different size and quality.} 
%To determine the optimal corpus size to build DLMs we extract DLMs from different large corpora. 
To evaluate the influence of the size and quality of the input corpus for building the DLMs, we experiment with corpora of different size and quality. 

We first evaluate with DLMs extracted from the different number of single-parsed sentences. We extracted DLMs start from a 5 million sentences corpus and increase the size of the corpus in step until all of the auto-parsed sentences are used. Table \ref{table-size} shows our results on English and Chinese development sets. For English, the highest accuracy is still achieved by DLM extracted from 5 million sentences. While for Chinese, we gain the largest improvement of 1.2\% with DLMs extracted from 10 million sentences. 

We further evaluate the influence of DLMs extracted from higher quality data. The higher quality corpora are prepared by parsing unlabeled sentences with the Mate parser and the Berkeley parser and adding the sentences to the corpus where both parsers agree. For Chinese, only 1 million sentences that consist of 5 tokens in average had the same syntactic structures assigned by the two parsers. Unfortunately, this amount is 
not sufficient for the experiments as their average sentence length is in stark contrast with the training data (27.1 tokens).
For English, we obtained 7 million sentences with an average sentence length of 16.9 tokens.

To get a first impression of the quality, we parsed the development set with the two parsers. When the parsers agree, the parse trees have an accuracy of 97\% LAS, while the labeled scores of both parsers are around 91\%. 
This indicates that parse trees where both parsers return the same tree have a higher accuracy.  The DLM extracted from 7 million higher quality English sentences achieved a higher accuracy of 91.56\%  which outperform the baseline by 0.51\%.

\begin{table}[t]
\begin{center}
\begin{adjustbox}{max width=7cm}
\begin{tabular}{l|l|l|l|l|l|l|l}
\hline \bf $m$ &\bf 0&\bf 1 &\bf   2&\bf 3&\bf 1-2&\bf  1-3 &\bf  1-4 \\ \hline
English&91.05 &\bf 91.43& 91.14& 91.22&91.27&91.26&N/A\\ 
Chinese&78.95 & 79.85&79.42& 79.06&79.97&\bf 80.11&79.73\\
 \hline

\end{tabular}
\end{adjustbox}
\end{center}
\caption{\label{table-order} Effects (LAS) of different number of DLMs for English and Chinese. $m$ = 0 refers the baseline.}
\end{table} 

\begin{table}[t]
\begin{center}
\begin{adjustbox}{max width=6.5cm}
\begin{tabular}{l|l|l|l|l|l}
\hline \bf Size  &\bf 0&\bf 5&\bf 10&\bf 20&\bf 30 \\ \hline
English&91.05& \bf 91.43&91.38&91.13&91.28\\
Chinese&78.95& 80.11&\bf80.15&79.72&N/A\\
 \hline

\end{tabular}
\end{adjustbox}
\end{center}
\caption{\label{table-size} Effects (LAS) of DLMs extracted from different size (in million sentences) of corpus. Size = 0 refers the baseline.}
\end{table}

\begin{table}[t]
\begin{center}
\begin{adjustbox}{max width=7cm}
\begin{tabular}{l|l|l|l|l}
\hline \bf System &\bf Beam&\bf POS &\bf LAS&\bf UAS \\ \hline
\newcite {zhang11}&32& 97.44&90.95&93.00\\ 
\newcite {bohnet2012:EACL2012}&80& 97.44&91.19&93.27\\ 
\newcite{martins2013tubo}&N/A& 97.44&90.55&92.89\\
\newcite{zhang-mcdonald:2014:P14-2}&N/A& 97.44&91.02&93.22\\ \hline
\newcite{chen2014fast}$\dagger$&1&N/A&89.60&91.80\\ 
\newcite{dyer-EtAl:2015:ACL-IJCNLP}$\dagger$&1&97.30&90.90&93.10\\ 
\newcite{weiss-etAl:2015:ACL}$\dagger$&8& 97.44&92.05&93.99\\ 
\newcite{andor-EtAl:2016:ACL}$\dagger$&32&97.44&92.79&94.61\\
\newcite{dozat2017deep}$\dagger$&N/A&N/A&94.08&95.74\\
\newcite{2017arXiv170705000L}$\dagger$&N/A&N/A&\bf95.20&\bf96.20\\
\hline
\newcite{chen2012utilizing} Baseline *&8&N/A&N/A&92.10\\
\newcite{chen2012utilizing} DLM *&8&N/A&N/A&92.76\\
Our Baseline *&40&97.33&92.44&93.38\\
\hline
Our Baseline&40&97.36&90.95&93.08\\
&80&97.34&91.05&93.28\\
&150&97.34&91.05&93.29\\ \hline
 Our DLM&40&97.38&91.41&93.59\\ 
 &80&97.39&91.47&93.65\\
 &150&97.42&91.56&93.74\\
 \hline

\end{tabular}
\end{adjustbox}
\end{center}
\caption{\label{table-encompare} Comparing with top performing parsers on English.  (* means results that are evaluated on \newcite{yamada03} conversion. $\dagger$ means neural network-based parsers)}
\end{table} 

\textbf{Main Results on Test Sets.}
We applied the best settings tuned on the development sets to the test sets. The best setting for English is the unigram DLM derived from the double parsed sentences. Table \ref{table-encompare} presents our results and top performing dependency parsers which were evaluated on the same English data set. Our approach with 40 beams surpasses our baseline by 0.46/0.51\% (LAS/UAS) \footnote{Significant in Dan Bikel's test ($p<10^{-3}$).} and is only lower than the few best neural network systems. When we enlarge the beam, our enhanced models achieved similar improvements. Our semi-supervised result with 150 beams are more competitive when compared with the state-of-the-art. We cannot directly compare our results with that of \newcite{chen2012utilizing} as they evaluated on an old \newcite{yamada03} format. In order to have an idea of the accuracy difference between our baseline and the second-order graph-based parser they used, we include our baseline on \newcite{yamada03} conversion. As shown in table \ref{table-encompare} our baseline is 0.62\% higher than their semi-supervised result and this is 1.28\% higher than their baseline. This confirms our claim that our baseline is much stronger.

For Chinese, we extracting the DLMs from 10 million sentences parsed by the Mate parser and using the unigram, bigram and the trigram DLMs together. Table \ref{table-cncompare} shows the results of our approach and a number of best Chinese parsers. Our system gained a large improvement of 0.93/0.98\%  \footnote{Significant in Dan Bikel's test ($p<10^{-5}$).} for labeled and unlabeled attachment scores when using a beam of 40. When larger beams are used our approach achieved even larger improvement of more than one percentage point for both labeled and unlabeled accuracy when compared to the respective baselines. Our scores with the default beam size (40) are competitive and are 0.2\% higher than the best reported result \cite{chen2013feature} when increasing the beam size to 150. 
Moreover, we gained improvements up to 0.42\% for part-of-speech tagging on Chinese tests.

%All improvements are significant in Dan Bikel's test with the default setting of 10,000 iterations ($p<10^{-3}$).

%The improvements are tested by Dan Bikel's randomized parsing evaluation comparator with the default settings of 10,000 iterations. All improvements are significant in Dan Bikel's test ($p<10^{-3}$).

\begin{table}[t]
\begin{center}
\begin{adjustbox}{max width=6.5cm}
\begin{tabular}{l|l|l|l|l}
\hline \bf System &\bf Beam&\bf POS&\bf LAS&\bf UAS \\ \hline
\newcite{hatori-EtAl:2011:IJCNLP-2011}&64&93.94&N/A&81.33\\ 
\newcite{li-EtAl:2012:PAPERS4}&N/A&94.60&79.01&81.67\\ 
\newcite{chen2013feature}&N/A&N/A&N/A&83.08\\ 
\newcite{chen2015feature}&N/A&93.61&N/A&82.94\\ \hline
Our Baseline&40&93.99&78.49&81.52\\
&80&94.02&78.48&81.58\\
&150&93.98&78.96&82.11\\ \hline
Our DLM&40&94.27&79.42&82.51\\
 &80&94.39&79.79&82.79\\
 &150&94.40&\bf80.21&\bf83.28\\
 \hline
\end{tabular}
\end{adjustbox}
\end{center}
\caption{\label{table-cncompare} Comparing with top performing parsers on Chinese.}
\end{table}

\section{Conclusion}
\label{conclusion}
In this paper, we applied dependency language models (DLM) extracted from a large parsed corpus to a strong transition-based parser. 
We integrated a small number of DLM-based features into the parser. 
We demonstrate the effectiveness of our DLM-based approach by applying our approach to English and Chinese. 
%We introduced additionally a data selection step to train our parser with DLMs derived by high quality source data. As a result, 
We achieved statistically significant improvements on labeled and unlabeled scores of both languages. Our parsing system improved by DLMs outperforms most of the systems on English and is competitive. For Chinese, we gained a large improvement of one point and our accuracy is 0.2\% higher than the best reported result. In addition to that, our approach gained an improvement of 0.4\% on Chinese part-of-speech tagging.

%\section*{Acknowledgments}

%The acknowledgments should go immediately before the references.  Do
%not number the acknowledgments section. Do not include this section
%when submitting your paper for review.

\bibliographystyle{acl_natbib}
\bibliography{acl2017,main2}

\end{document}